\documentclass[twoside,11pt]{article}

\usepackage{blindtext}

\usepackage{jmlr2e}
\usepackage{times}  
\usepackage{helvet}  
\usepackage{courier}  
\usepackage{natbib}  
\usepackage{caption} 
\usepackage{newfloat}
\usepackage{listings}
\usepackage{xcolor}
\usepackage[most]{tcolorbox}
\usepackage{lipsum}
\usepackage{algorithm}
\usepackage{algorithmic}
\usepackage{booktabs}

\usepackage{lastpage}
\jmlrheading{25}{2025}{1-\pageref{LastPage}}{8/25}{--/--}{--}{Si Shen, Peijun Shen, Wenhua Zhao and Danhao Zhu}

\ShortHeadings{S-GRPO}{Shen, Shen, Zhao, and Zhu}
\firstpageno{1}

\begin{document}

\title{Mitigating Think-Answer Mismatch in LLM Reasoning Through Noise-Aware Advantage Reweighting}

\author{\name Si Shen \email shensi@njust.edu.cn \\
       \addr Department of Computer Science and Engineering\\
       Nanjing University of Science and Technology\\
       Nanjing, 210094, China
       \AND
       \name Peijun Shen \email spj020212@njust.edu.cn \\
       \addr Department of Computer Science and Engineering\\
       Nanjing University of Science and Technology\\
       Nanjing, 210094, China
       \AND
       \name Wenhua Zhao \email zhaowenhua@njau.edu.cn \\
       \addr Department of Information Management \\
       Nanjing Agricultural University \\
       Nanjing, 210095, China
       \AND
       \name Danhao Zhu\thanks{Corresponding Author} \email zhudanhao@jspi.edu.cn \\
       \addr Department of Criminal Science and Technology \\
       Jiangsu Police Institute \\
       Nanjing, 210031, China}

\editor{My editor}

\maketitle

\begin{abstract}
Group-Relative Policy Optimization (GRPO) is a key technique for training large reasoning models, yet it suffers from a critical vulnerability: the \emph{Think-Answer Mismatch}, where noisy reward signals corrupt the learning process. This problem is most severe in unbalanced response groups, paradoxically degrading the signal precisely when it should be most informative. To address this challenge, we propose Stable Group-Relative Policy Optimization (S-GRPO), a principled enhancement that derives optimal, noise-aware advantage weights to stabilize training. Our comprehensive experiments on mathematical reasoning benchmarks demonstrate S-GRPO's effectiveness and robustness. On various models, S-GRPO significantly outperforms DR. GRPO, achieving performance gains of +2.5\% on Qwen-Math-7B-Base, +2.2\% on Llama-3.2-3B-Base, and +2.4\% on Qwen-Math-1.5B-Instruct. Most critically, while standard GRPO fails to learn under 20\% synthetic reward noise, S-GRPO maintains stable learning progress. These results highlight S-GRPO's potential for more robust and effective training of large-scale reasoning models. \footnote{Code and data are available at: \url{https://github.com/shenpeijun0212/S-GRPO}}
\end{abstract}


%

\section{Introduction}
Recent breakthroughs in large-scale reasoning models are largely attributed to methods like Group-Relative Policy Optimization (GRPO)~\citep{shao2024deepseekmath}, a reinforcement learning technique that has propelled models such as DeepSeek-R1 and Qwen3 to state-of-the-art performance on challenging reasoning benchmarks \citep{guo2025deepseek, shao2024deepseekmath, liu2025understanding}. GRPO's efficacy stems from a simple yet powerful principle: rewarding responses that yield correct final answers relative to a group of sampled outputs, while penalizing incorrect ones. This approach obviates the need for an explicit value function, significantly simplifying the training pipeline.

Despite its empirical success, GRPO's reliance on final-answer correctness as a proxy for reasoning quality presents a critical vulnerability. A correct answer does not necessarily imply valid or logically sound reasoning \citep{tyen2023llms, zheng2024processbench, song2025prmbench}. For instance, \citet{zheng2024processbench} report that Qwen and LLaMA models exhibit reasoning error rates ranging from 3.5\% to 51.8\% across various benchmarks, even when their final answers are correct. Conversely, an incorrect final answer does not always indicate flawed reasoning. As shown by \citet{kiciman2023causal}, models like GPT-3.5 can produce intermediate reasoning steps that successfully identify and correct earlier mistakes, yet ultimately revert to an incorrect final answer. This phenomenon, commonly referred to as the \emph{Think-Answer Mismatch} \citep{yao2025are, chen2025reasoning}, has been consistently observed across diverse models, tasks, and evaluation protocols.

\begin{figure}[t]
	\centering
	\includegraphics[width=0.80\textwidth]{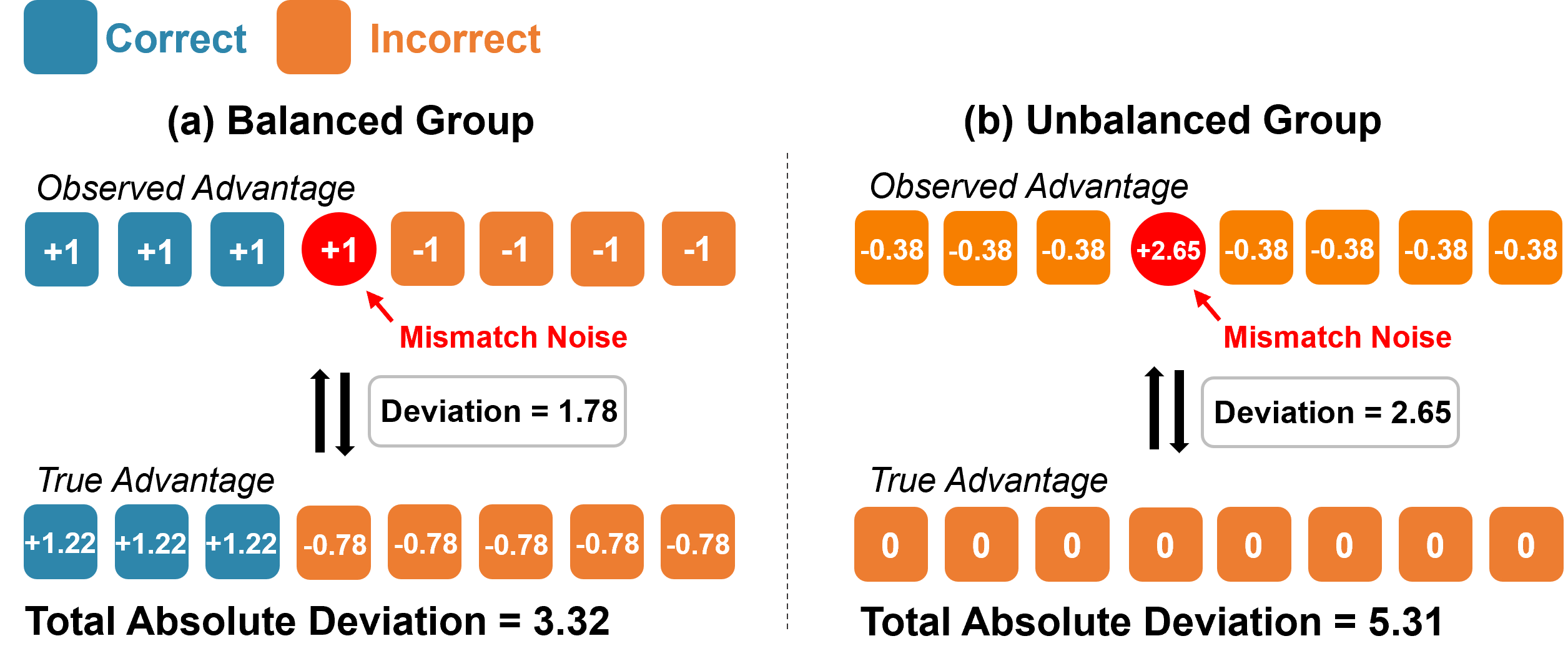}
	\caption{The impact of a single Think-Answer Mismatch on GRPO's advantage calculation in a balanced group (4 correct, 4 incorrect responses) versus an unbalanced group (1 correct, 7 incorrect responses). A false positive, where flawed reasoning leads to a correct answer (indicated by the red circle), causes a much larger deviation in the unbalanced group.}
	\label{fig:main-illustration}
\end{figure}

The consequences of this misalignment are particularly pronounced for GRPO. Our analysis uncovers a specific failure mode: vulnerability to reward noise in unbalanced groups. As illustrated in Figure~\ref{fig:main-illustration}, a single false positive mismatch, where a response with flawed reasoning that happens to yield the correct answer, can have a disproportionately large impact depending on group composition. In a highly unbalanced group (e.g., one correct answer out of eight), this single mismatch sample can severely distort the advantage signal, inflating the overall observed advantage by up to 60\% (5.31 vs. 3.32) compared to a balanced group. This creates a paradox: precisely when the learning signal should be strongest, that is, when a rare success occurs among many failures, it becomes most vulnerable to corruption. As demonstrated in Section 2, this vulnerability is not merely theoretical; under noise levels (e.g., 20\%), the standard GRPO learning process can collapse entirely.

To address this critical vulnerability, we propose Stable Group-Relative Policy Optimization (S-GRPO), a principled enhancement that explicitly models and mitigates the impact of reward noise. Our approach is founded on the insight that balanced groups provide an inherently more robust training signal. S-GRPO operationalizes this by deriving an optimal, closed-form advantage weight that minimizes the expected squared error between the observed and true advantages under a symmetric noise model. This reweighting scheme automatically down-weights signals from unbalanced groups where noise has an outsized impact, yielding a method that maintains the computational efficiency of GRPO while gracefully handling noisy rewards.

Our comprehensive evaluations demonstrate that S-GRPO consistently outperforms standard GRPO under identical experimental setups, achieving significant gains on models like Qwen-Math-7B (+2.5\%), Llama-3.2-3B (+2.2\%), and Qwen-Math-1.5B (+2.4\%). More importantly, S-GRPO demonstrates remarkable robustness: while standard GRPO fails to learn under 20\% synthetic label noise, S-GRPO maintains stable training progress with minimal performance degradation. Further analysis reveals that S-GRPO fosters a more stable and efficient training process, evidenced by smoother entropy reduction and the emergence of more reliable reasoning patterns.

Our contributions are three-fold:
\begin{itemize}
	\item We are the first to identify and formalize the vulnerability of GRPO to the \emph{Think-Answer Mismatch}, showing how its impact is amplified by group imbalance.
	\item We propose S-GRPO, a principled extension that derives optimal, noise-aware advantage weights to ensure robust policy updates.
	\item We demonstrate empirically that S-GRPO improves performance, robustness, and training stability across multiple reasoning benchmarks and model scales.
\end{itemize}

\section{Robustness of GRPO to Reward Noise}

\subsection{Background: Group-Relative Policy Optimization}

Group-Relative Policy Optimization (GRPO) is a memory-efficient reinforcement learning algorithm designed for fine-tuning Large Language Models (LLMs). Its core innovation lies in eliminating the need for a separate critic model, a staple in traditional Reinforcement Learning from Human Feedback (RLHF) methods like Proximal Policy Optimization (PPO)\citep{schulman2017proximal}. Instead, GRPO computes the advantage for each response relative to a baseline derived from a group of peer responses generated for the same query, significantly reducing computational overhead during training.

For a given input query $q$, the actor LLM generates a group of $N$ responses, $\{o_i\}_{i=1}^N$. Each response is assigned a binary reward $r_i \in \{0, 1\}$, indicating whether it yields the correct final answer. The original GRPO framework defines the advantage by standardizing this reward within the group:
\begin{equation}
	a_i = \frac{r_i - \bar{r}}{\sqrt{\bar{r}(1 - \bar{r}) + \epsilon}},
	\label{eq:grpo_advantage}
\end{equation}
where $\bar{r} = \frac{1}{N}\sum_{i=1}^N r_i$ is the empirical mean reward of the group and $\epsilon$ is a small constant to prevent division by zero. This group-wise normalization centers the advantages around zero and scales them to unit variance, which helps stabilize the learning process.

\subsection{The Impact of Think-Answer Mismatch on Advantage Calculation}

We now analyze how a 'Think-Answer Mismatch' false positive, where reasoning is flawed but the final answer is correct, impacts advantage computation in GRPO. The impact of false negatives remains the same. Consider a group of $N$ responses where $k$ responses are observed to have a reward of 1. Under the original GRPO formulation, the observed positive advantage ($a_{\text{pos}}$) and negative advantage ($a_{\text{neg}}$) are:
\begin{equation}
	\begin{split}
		a_{\text{pos}} = \frac{N-k}{\sqrt{k(N-k)}}   \\
		a_{\text{neg}} = \frac{-k}{\sqrt{k(N-k)}}
	\end{split}
\end{equation}

Now consider the case where one of these observed positive rewards is actually incorrect due to a Think-Answer Mismatch. The true reward distribution should have $(k-1)$ positive and $(N-k+1)$ negative responses. The corrected advantages become:
\begin{equation}
	\begin{split}
		a_{\text{pos}}^{\text{true}} &= \frac{N-k+1}{\sqrt{(k-1)(N-k+1)}} \\
		a_{\text{neg}}^{\text{true}} &= \frac{-(k-1)}{\sqrt{(k-1)(N-k+1)}}
	\end{split}
\end{equation}

The total absolute deviation in advantages across the entire group can be formulated as:
\begin{align}
	\Delta_{\text{total}} &= \underbrace{\left| a_{\text{pos}} - a_{\text{neg}}^{\text{true}} \right|}_{\text{Mismatch Sample}} \nonumber \\
	&\quad + \underbrace{(k-1) \times \left| a_{\text{pos}} - a_{\text{pos}}^{\text{true}} \right|}_{\text{True Positives}} \nonumber \\
	&\quad + \underbrace{(N-k) \times \left| a_{\text{neg}} - a_{\text{neg}}^{\text{true}} \right|}_{\text{True Negatives}}
	\label{eq:total_deviation}
\end{align}

\begin{figure}[h]
	\centering
	\includegraphics[width=0.50\textwidth]{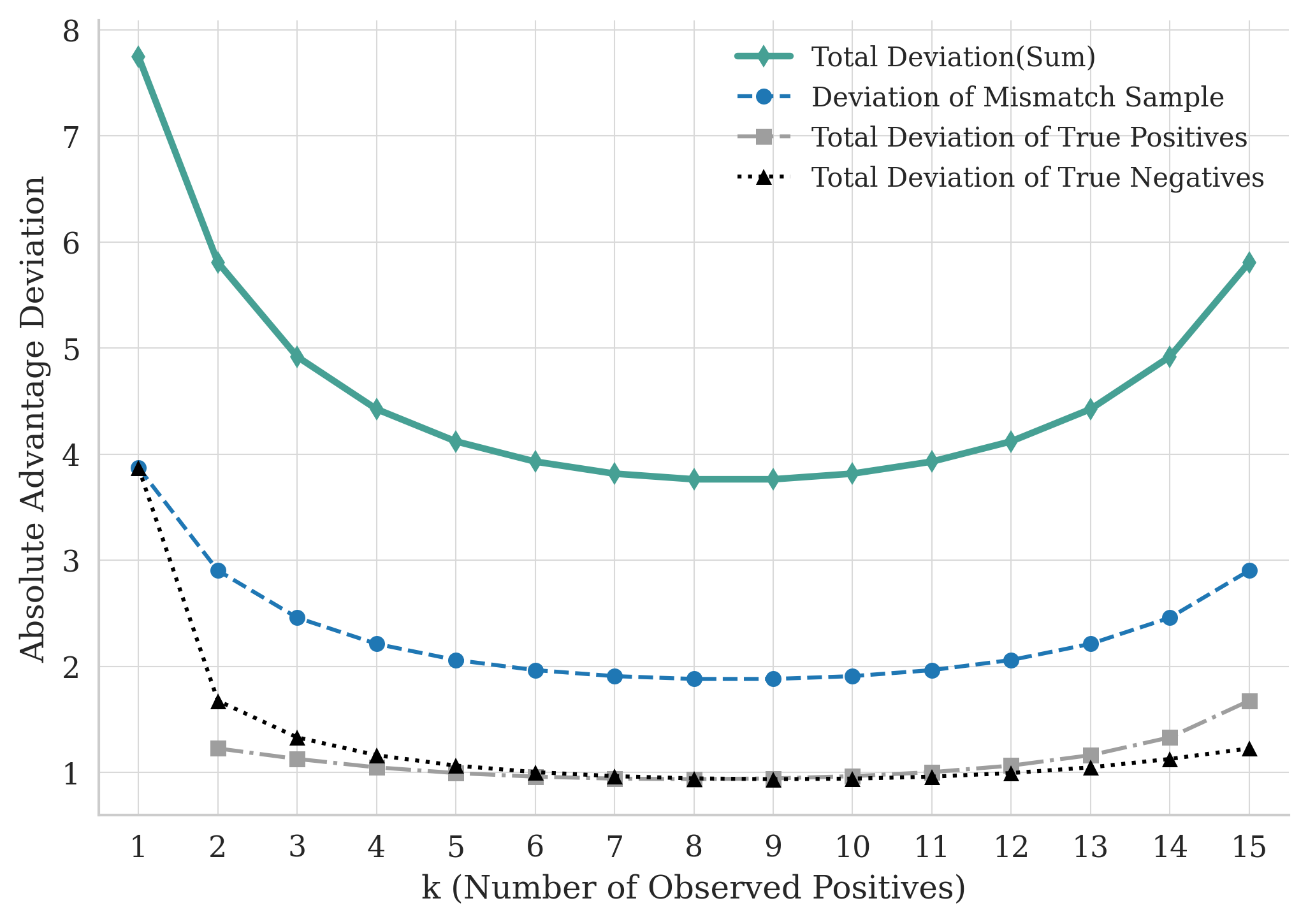} 
	\caption{The impact of a single false positive Think-Answer Mismatch on GRPO's advantage calculation in groups of size $N = 16$.}
	\label{fig:advantage_deviation}
\end{figure}

This formulation reveals that a single Think-Answer Mismatch creates a cascading error: not only does the mismatch sample receive an incorrect advantage signal, but all other samples in the group also experience advantage distortions due to the altered group statistics. Figure~\ref{fig:advantage_deviation} illustrates this effect for a group size of $N=16$ ($N=16$ here for clarity and all other experiments use $N=8$). The deviation exhibits a U-shaped relationship with group composition, being most pronounced in imbalanced groups.

\subsection{Consequences for Training Dynamics}

\begin{figure}[h]
	\centering
	\includegraphics[width=0.80\textwidth]{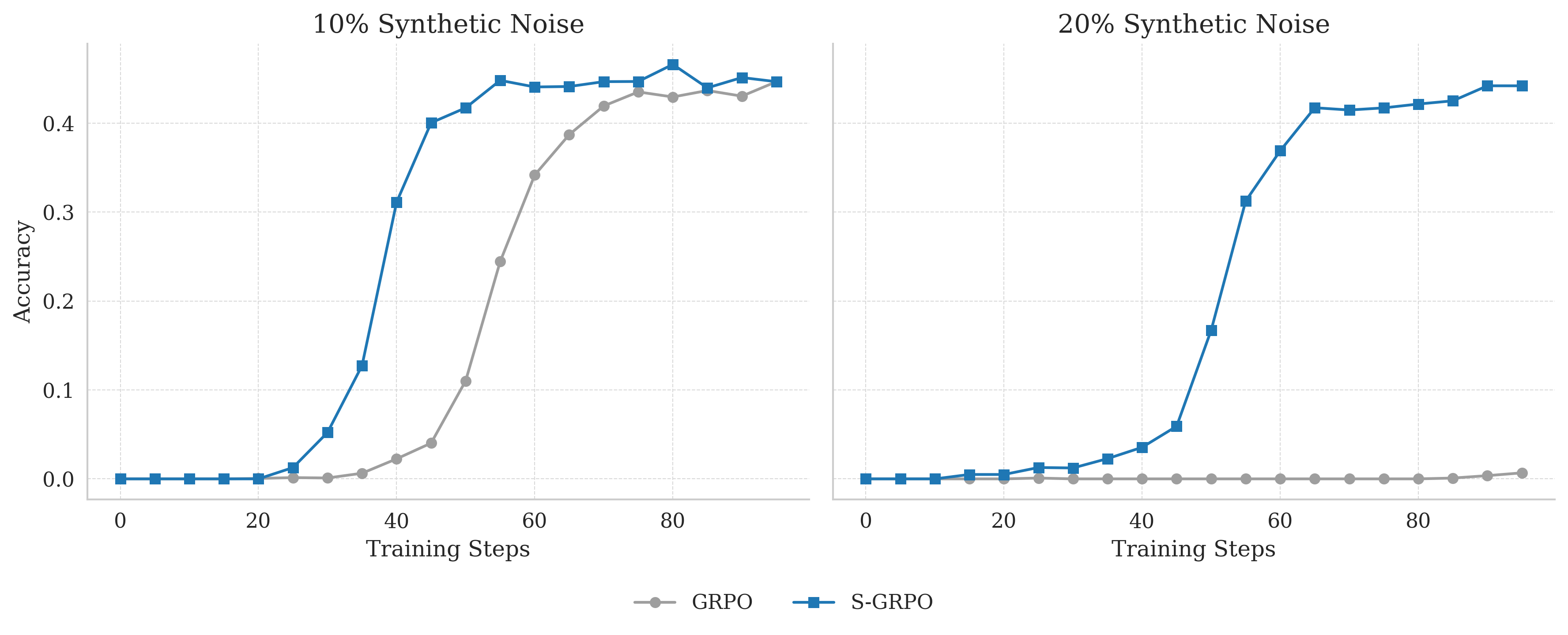}
	\caption{Impact of synthetic reward noise on the training dynamics of S-GRPO and standard GRPO. Pass@1 accuracy over 100 training steps for GRPO and S-GRPO under synthetic noise levels of 10\% (left) and 20\% (right).}
	\label{fig:training_dynamics}
\end{figure}

To investigate how Think-Answer Mismatches affect training, we conduct experiments where we inject synthetic noise into the reward signal. Specifically, we randomly flip the binary reward of a response, simulating an increased rate of mismatches. Figure~\ref{fig:training_dynamics} shows the results for noise levels of 10\% and 20\%.

With 10\% injected noise, standard GRPO exhibits significantly slower learning compared to our proposed S-GRPO. When the noise level increases to 20\%, GRPO's performance collapses entirely, failing to show any meaningful learning within the first 100 steps. In contrast, S-GRPO continues to learn effectively under both noise levels. These results demonstrate that as the rate of Think-Answer Mismatches increases, standard GRPO's performance degrades substantially, while S-GRPO maintains robust learning capabilities.

\section{S-GRPO: Stable GRPO Through Noise-Aware Reweighting}

To mitigate the adverse effects of Think-Answer Mismatches, we introduce S-GRPO (Stable GRPO), which denoises the advantage signal through principled reweighting based on an explicit noise model.

\subsection{Symmetric Reward Noise Model}
We model the Think-Answer Mismatch as symmetric label noise~\cite{angluin1988learning,van2015learning}. Each observed reward $r_i$ is an independent flip of the latent true reward $r_i^* \in \{0,1\}$ with a fixed probability $p$:  
\begin{equation}
	\mathbb{P}(r_i \ne r_i^*) = p, \quad \text{where } 0 \le p < 0.5.
\end{equation}  
Here, $p$ is a hyperparameter representing the probability of a Think-Answer Mismatch. In practice, this value tends to be higher for more complex datasets and weaker models, and lower for simpler datasets and stronger models.

Given the observed mean reward $\bar{r} = k/N$ for a group of $N$ responses, the expected true mean reward $t = \mathbb{E}[r_i^*]$ can be estimated as:
\begin{equation}
	t = \frac{\bar{r} - p}{1 - 2p}.
\end{equation}

This relationship follows from $\bar{r} = \mathbb{P}(r_i=1) = (1-p)t + p(1-t)$. We clip $t$ to $[0, 1]$ to ensure validity. Intuitively, if the observed success rate $\bar{r}$ approaches the noise rate $p$, the estimated true success rate $t$ approaches 0.

\subsection{Denoised Advantage via Optimal Reweighting}

Our goal is to find an optimal weight $w^{\star}$ for each group that minimizes the expected squared error between the reweighted observed advantage and the unobserved true advantage. The standardized advantages are:
\begin{equation}
	a_i = \frac{r_i - \bar{r}}{\sigma_r} \quad \text{and} \quad a_i^* = \frac{r_i^* - t}{\sigma_t},
\end{equation}
where $\sigma_r^2 = \bar{r}(1 - \bar{r}) + \epsilon$ and $\sigma_t^2 = t(1 - t) + \epsilon$.

We seek to solve:
\begin{equation}
	w^\star = \arg\min_w \mathcal{L}(w) = \mathbb{E}\left[(w a_i - a_i^*)^2\right].
	\label{eq:loss_function}
\end{equation}

Since both $a_i$ and $a_i^*$ are standardized with zero mean and unit variance, expanding the loss function yields:
\begin{equation}
	\mathcal{L}(w) = w^2 - 2w \operatorname{Cov}(a_i, a_i^*) + 1.
\end{equation}

Setting the derivative to zero gives the optimal weight:
\begin{equation}
	w^{\star} = \operatorname{Cov}(a_i, a_i^*) = \frac{\operatorname{Cov}(r_i, r_i^*)}{\sigma_r \sigma_t}.
\end{equation}

The covariance between observed and true rewards is $\operatorname{Cov}(r_i, r_i^*) = (1-2p)t(1-t)$. Substituting yields:
\begin{equation}
	w^{\star}(N,k,p) = \frac{(1-2p)t(1-t)}{\sqrt{\bar{r}(1-\bar{r})+\epsilon}\sqrt{t(1-t)+\epsilon}}.
	\label{eq:optimal_w_final}
\end{equation}

This weight represents the correlation coefficient between observed and true rewards, scaled by the noise factor $(1-2p)$. S-GRPO uses the reweighted advantage $w^\star a_i$ for policy gradient updates.

\subsubsection{Analysis of the Optimal Weight}

The behavior of $w^{\star}$ (Equation 11) reveals several key properties aligned with our goal of robust learning:

\begin{figure}[h]
	\centering
	\includegraphics[width=0.60\textwidth]{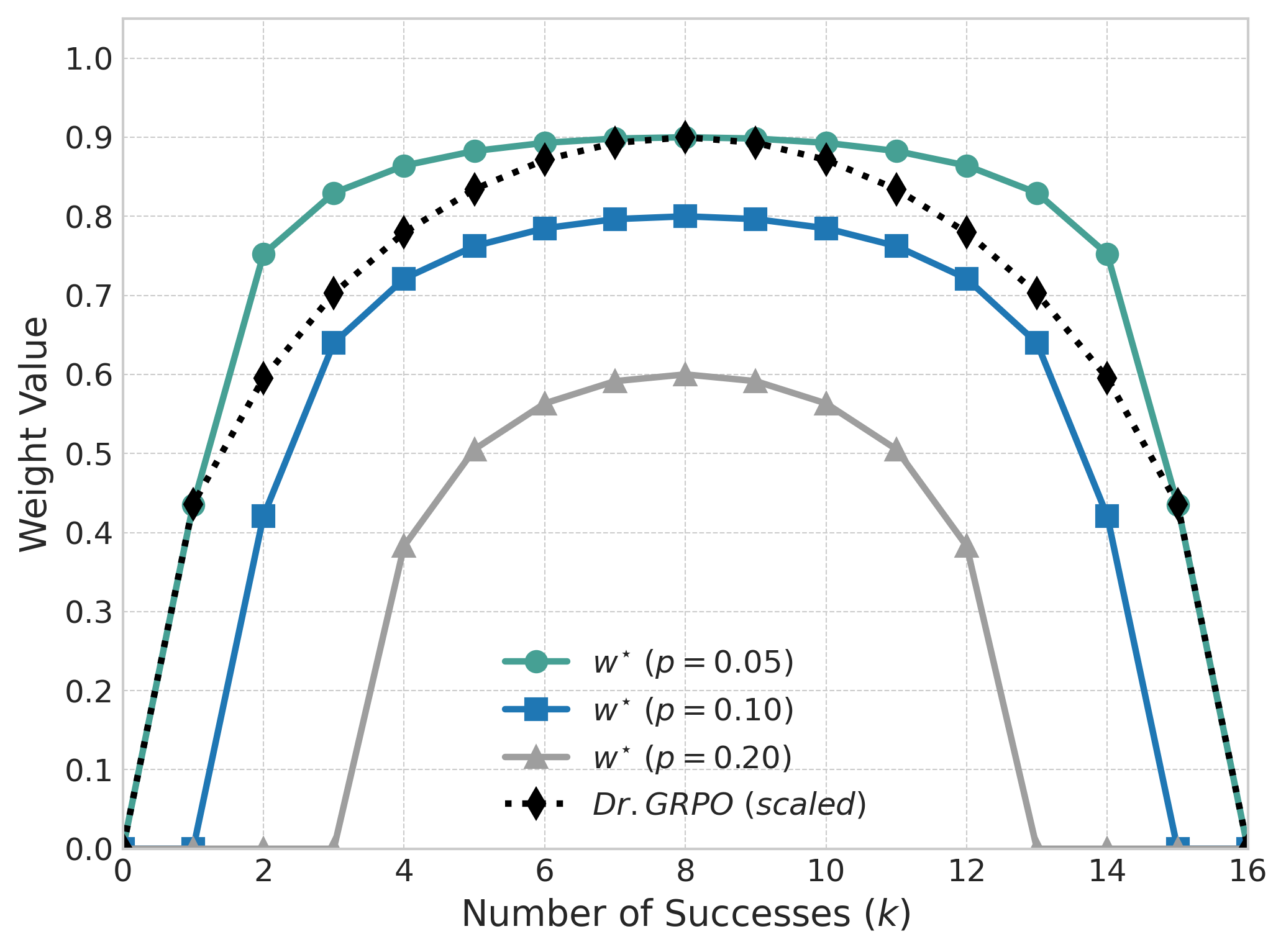}
	\caption{The optimal weight $w$ as a function of successful responses $k$ in a group of size $N = 16$ for different assumption noise levels $p$. Dr. GRPO's reweighting strategy (scaled to maximum $0.9$) is shown for comparison.}
	\label{fig:weight_analysis}
\end{figure}

As illustrated in Figure~\ref{fig:weight_analysis}, the weight $w^{\star}$ exhibits three important characteristics:

\begin{enumerate}
	\item \textbf{Noise-Adaptive Attenuation:} The weight is bounded by $(1-2p)$, with higher noise levels uniformly down-weighting the learning signal. In the noiseless case ($p=0$), $w^{\star}=1$, recovering the original GRPO.
	
	\item \textbf{Confidence through Consensus:} The weight is smallest for highly imbalanced groups and largest for balanced ones ($k \approx N/2$). This concave shape formalizes the intuition that balanced groups provide more reliable signals.
	
	\item \textbf{Noise-Gating Mechanism:} When the observed success rate falls below the assumed noise rate $p$, the weight becomes zero. For example, with $p=0.20$, groups with $k \le 3$ or $k \geq 13$ (out of 16) are completely gated, preventing updates based on statistically unreliable signals.
\end{enumerate}

\paragraph{Comparison with Existing Methods.}
Figure~\ref{fig:weight_analysis} also shows Dr.~GRPO's heuristic \citep{liu2025understanding}, which removes standard deviation normalization. While both approaches upweight balanced groups, Dr.~GRPO lacks noise adaptivity and the hard gating mechanism for low-confidence cases, properties that prior heuristics like DAPO \citep{yu2025dapo} and Seed-GRPO \citep{chen2025seed} also cannot provide.

\subsection{Optimization Objective}

We adopt a clipped surrogate objective inspired by PPO, adapted for noise-aware reweighting:
\begin{equation}
	\begin{split}
		\mathcal{L}_i(\theta) = \min \big( &\text{ratio}_i(\theta) w^\star a_i, \\
		&\text{clip}(\text{ratio}_i(\theta), 1-\epsilon, 1+\epsilon) w^\star a_i \big),
	\end{split}
\end{equation}
where $\text{ratio}_i(\theta) = \frac{\pi_\theta(o_i \mid q)}{\pi_{\theta_{\text{old}}}(o_i \mid q)}$ is the importance sampling ratio.

The overall training objective for query $q$ with $G$ responses is:
\begin{equation}
	\mathcal{L}(\theta) = \frac{1}{G} \sum_{i=1}^G \mathcal{L}_i(\theta).
\end{equation}

Model parameters are updated via stochastic gradient ascent following the standard PPO scheme.

\section{Experiments}

In this section, we empirically evaluate S-GRPO across multiple dimensions. We investigate the following research questions:
\begin{itemize}
	\item \textbf{RQ1}: How does S-GRPO compare to baselines on mathematical reasoning benchmarks? 
	\item \textbf{RQ2}: How do different noise assumptions (p values) affect S-GRPO's performance and training stability?
	\item \textbf{RQ3}: What emergent behaviors and characteristics does S-GRPO exhibit compared to standard GRPO?
\end{itemize}

\subsection{Experimental Setup}

\subsubsection{Datasets}
We conduct reinforcement learning on 8,500 problems sampled from the MATH dataset \citep{hendrycks2021measuring}, specifically selecting problems with difficulty levels 3-5 to ensure appropriate challenge for our models. For evaluation, we employ four standard mathematical reasoning benchmarks. AMC (83 problems), MATH500 (500 problems) , Minerva (272 problems)\citep{lewkowycz2022solving}, OlympiadBench (475 problems)\citep{huang2024olympicarena}. We exclude the commonly used AIME24 benchmark due to its limited size (30 problems), which leads to unstable results, particularly for smaller models\footnote{See \url{https://github.com/sail-sg/understand-r1-zero/issues/21} for discussion on AIME24's instability.}.

\subsubsection{Baselines}

We compare S-GRPO against several strong baselines representing the current state-of-the-art in mathematical reasoning. Among the advanced reasoning models, we include RAFT++ \citep{xiong2025minimalist}, OpenReasoner-Zero-7B \citep{hu2025open} and SimpleRL-Zoo-7B \citep{zeng2025simplerl}.

For the most direct comparison to our approach, we evaluate against: the original GRPO \citep{shao2024deepseekmath}, and Dr.~GRPO \citep{liu2025understanding}. These two baselines are particularly important as they share nearly identical experimental settings with S-GRPO, providing the fairest comparison for evaluating our noise-aware reweighting strategy.

\subsubsection{Evaluation Metrics}

We report Pass@1 accuracy as our primary metric across all benchmarks, using greedy sampling for deterministic evaluation. To address the inherent instability in RL training, we report the average of the top-3 checkpoint performances, evaluated every 16 training steps within a maximum of 500 steps. This approach provides a more robust assessment of model capabilities while accounting for training variance.

\subsubsection{Training Details}

We train S-GRPO on three diverse base models to ensure generalizability: Qwen2.5-Math-7B-Base, Qwen2.5-Math-1.5B-Instruct, and Llama-3.2-3B-Base. This selection covers different model scales from 1.5B to 7B parameters and includes both base and instruction-tuned variants, addressing concerns about method sensitivity to base model choice \citep{zuo2025ttrl,shao2025spurious}. Additional hyperparameters and optimization details are provided in Appendix A.

%
%

\subsection{Main Results (RQ1)}

\begin{table*}[t]
	\centering
	\begin{tabular}{lccccc}
		\toprule
		\textbf{Model} & \textbf{AMC} & \textbf{MATH500} & \textbf{Minerva} & \textbf{OlympiadBench} & \textbf{Average} \\
		\midrule
		\multicolumn{6}{l}{\textit{State-of-the-art reasoning models}} \\
		RAFT++ 7B & - & 80.5 & 35.8 & 41.2 & - \\
		OpenReasoner-Zero-7B & 47.0 & 79.2 & 31.6 & 44.0 & 50.5 \\
		SimpleRL-Zoo-7B & 60.2 & 78.2 & 27.6 & 40.3 & 51.6 \\
		\midrule
		\multicolumn{6}{l}{\textit{Qwen2.5-Math-1.5B-Instruct}} \\
		Base Model & 43.4 & 61.8 & 15.1 & 28.4 & 37.2 \\
		GRPO-1.5B$^\star$ & 47.0$_{\pm 2.4}$ & 74.0$_{\pm 0.4}$ & 23.2$_{\pm 0.4}$ & 39.3$_{\pm 0.6}$ & 46.3$_{\pm 0.4}$ \\
		Dr.~GRPO-1.5B$^\star$ & 48.2$_{\pm 2.4}$ & 75.8$_{\pm 0.6}$ & 25.0$_{\pm 0.7}$ & 40.1$_{\pm 0.7}$ & 47.3$_{\pm 0.5}$ \\
		\textbf{S-GRPO-1.5B}$^\star$ & \textbf{51.8}$_{\pm 1.2}$ & \textbf{77.8}$_{\pm 0.2}$ & \textbf{27.6}$_{\pm 0.4}$ & \textbf{42.2}$_{\pm 0.3}$ & \textbf{49.7}$_{\pm 0.3}$ \\
		\midrule
		\multicolumn{6}{l}{\textit{Llama-3.2-3B-Base}} \\
		Base Model & 2.4 & 6.4 & 6.3 & 1.3 & 3.3 \\
		GRPO-3B$^\star$ & \textbf{7.2}$_{\pm 1.2}$ & 12.2$_{\pm 0.8}$ & 10.3$_{\pm 1.1}$ & 3.5$_{\pm 0.6}$ & 8.3$_{\pm 0.4}$ \\
		Dr.~GRPO-3B & \textbf{7.2} & 10.0 & 11.0 & 2.2 & 7.6 \\
		\textbf{S-GRPO-3B}$^\star$ & \textbf{7.2}$_{\pm 1.2}$ & \textbf{14.2}$_{\pm 0.4}$ & \textbf{12.9}$_{\pm 0.7}$ & \textbf{4.8}$_{\pm 0.4}$ & \textbf{9.8}$_{\pm 0.4}$ \\
		\midrule
		\multicolumn{6}{l}{\textit{Qwen2.5-Math-7B-Base}} \\
		Base Model & 45.8 & 69.0 & 21.3 & 34.7 & 42.7 \\
		GRPO-7B$^\star$ & 57.8$_{\pm 3.6}$ & 79.2$_{\pm 1.8}$ & 29.4$_{\pm 1.5}$ & 41.5$_{\pm 2.1}$ & 51.5$_{\pm 1.3}$ \\
		Dr.~GRPO-7B & \textbf{62.7} & 80.0 & 30.1 & 41.0 & 53.5 \\
		\textbf{S-GRPO-7B}$^\star$ & 61.5$_{\pm 2.4}$ & \textbf{82.2}$_{\pm 1.4}$ & \textbf{35.7}$_{\pm 1.1}$ & \textbf{45.3}$_{\pm 1.5}$ & \textbf{56.0}$_{\pm 0.4}$ \\
		\bottomrule
	\end{tabular}
	\caption{Performance comparison across mathematical reasoning benchmarks. Results marked with $\star$ indicate experiments we conducted under identical settings. Other results are from original papers. We report mean $\pm$ standard deviation for the top-3 checkpoints. The best results from models with same size are in bold.}
	\label{tab:main_results}
\end{table*}

Table~\ref{tab:main_results} presents our main experimental results across all benchmarks and models.
Comparison with State-of-the-Art Methods. S-GRPO demonstrates strong performance against competitive baselines. Our 7B model achieves an average accuracy of 56.0\%, shows S-GRPO as a highly competitive approach for mathematical reasoning.

Since RL performance is highly sensitive to base models and hyperparameters, the most informative comparisons are with GRPO and Dr.~GRPO under identical experimental settings. S-GRPO consistently outperforms both baselines across all three base models. On Qwen2.5-Math-1.5B-Instruct, S-GRPO achieves 49.7\% average accuracy, representing a 2.4 percentage point improvement over Dr.~GRPO. Similar gains are observed on Llama-3.2-3B-Base (+2.2 points) and Qwen2.5-Math-7B-Base (+2.5 points). The consistency of these improvements across diverse model architectures and scales validates our theoretical analysis: noise-aware reweighting provides a principled solution to the Think-Answer Mismatch problem.

\subsection{Robustness to Noisy Rewards (RQ2)}
While Section 2.3 demonstrated S-GRPO's robustness under synthetic noise injection, here we examine how different noise assumption levels of $p$ affect training dynamics in practical settings. 

\subsubsection{Training Dynamics Under Different Noise Assumptions}

\begin{figure}[h]
	\centering
	\includegraphics[width=0.50\textwidth]{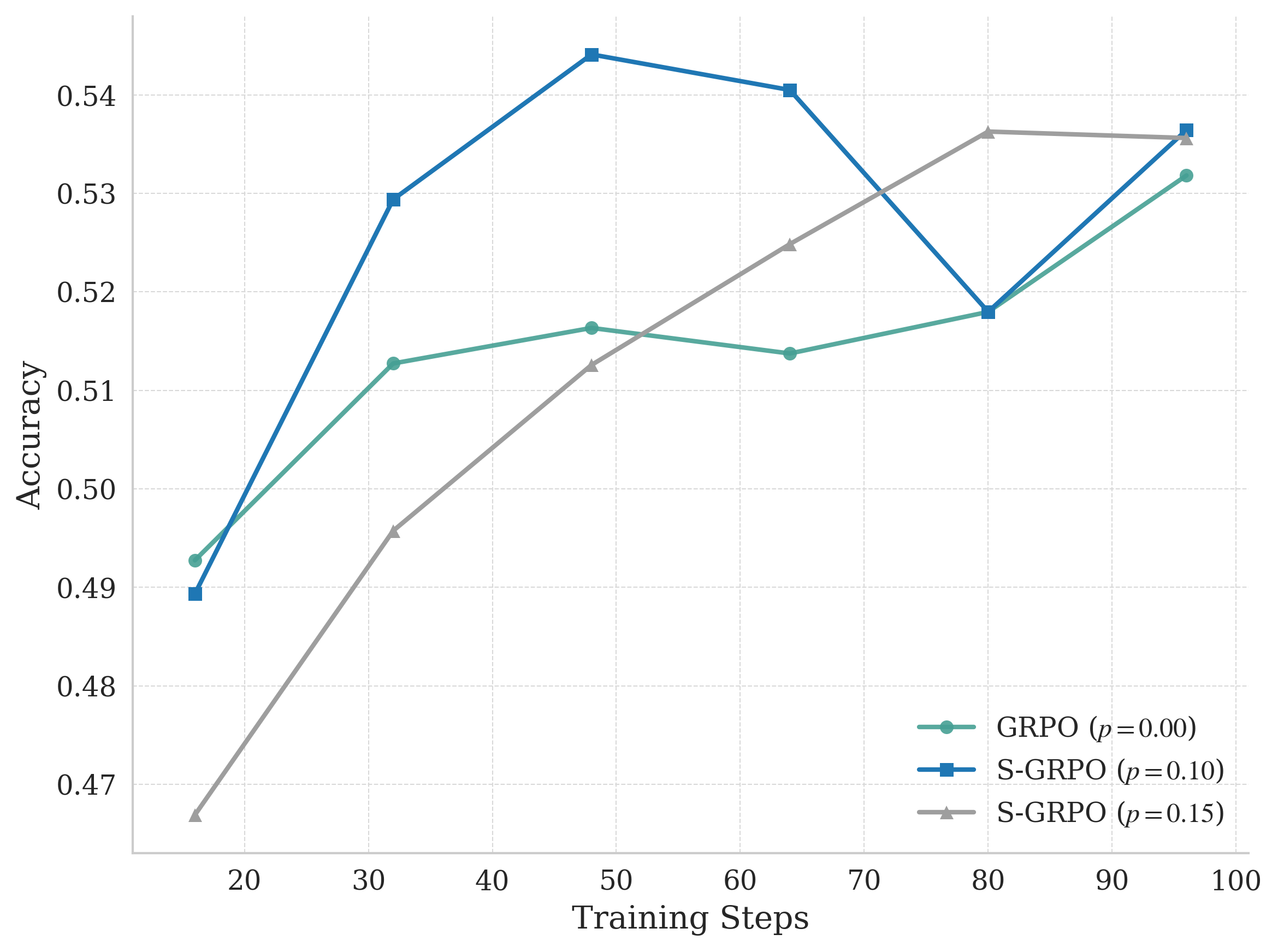}
	\caption{Training dynamics under different noise assumptions $p$ on Qwen2.5-Math-7B-Base. Higher noise assumptions lead to more stable long-term performance despite slower initial progress.}
	\label{fig:training_dynamics_noise}
\end{figure}

Figure~\ref{fig:training_dynamics_noise} illustrates S-GRPO's training dynamics with varying noise assumptions ($p = 0$, $0.10$, $0.15$) on Qwen2.5-Math-7B-Base during the first 100 training steps.

The results reveal a fundamental trade-off between learning speed and stability. With $p = 0.10$, the model achieves rapid initial improvement, jumping from 49\% to 54\% accuracy within 50 steps, significantly outperforming baseline GRPO which plateaus around 51-52\%. However, performance temporarily dips around step 80 before recovering. In contrast, $p = 0.15$ exhibits more conservative behavior: starting from a lower baseline (46\%), it shows steady, monotonic improvement throughout training, ultimately converging to similar performance levels by step 100.

This behavior stems from S-GRPO's noise-gating mechanism. At $p = 0.15$, groups with extreme imbalance ($k \in \{1, 7\}$ for $N = 8$) receive zero weight, effectively filtering out potentially misleading signals. While this initially slows learning, it prevents corrupted updates from groups where a single mismatch could dominate the advantage calculation. The resulting trade-off suggests practitioners should choose $p$ based on their requirements: lower values for rapid initial gains, higher values for monotonic improvement and long-term stability.

\subsubsection{Policy Entropy Evolution}

\begin{figure}[h]
	\centering
	\includegraphics[width=0.50\textwidth]{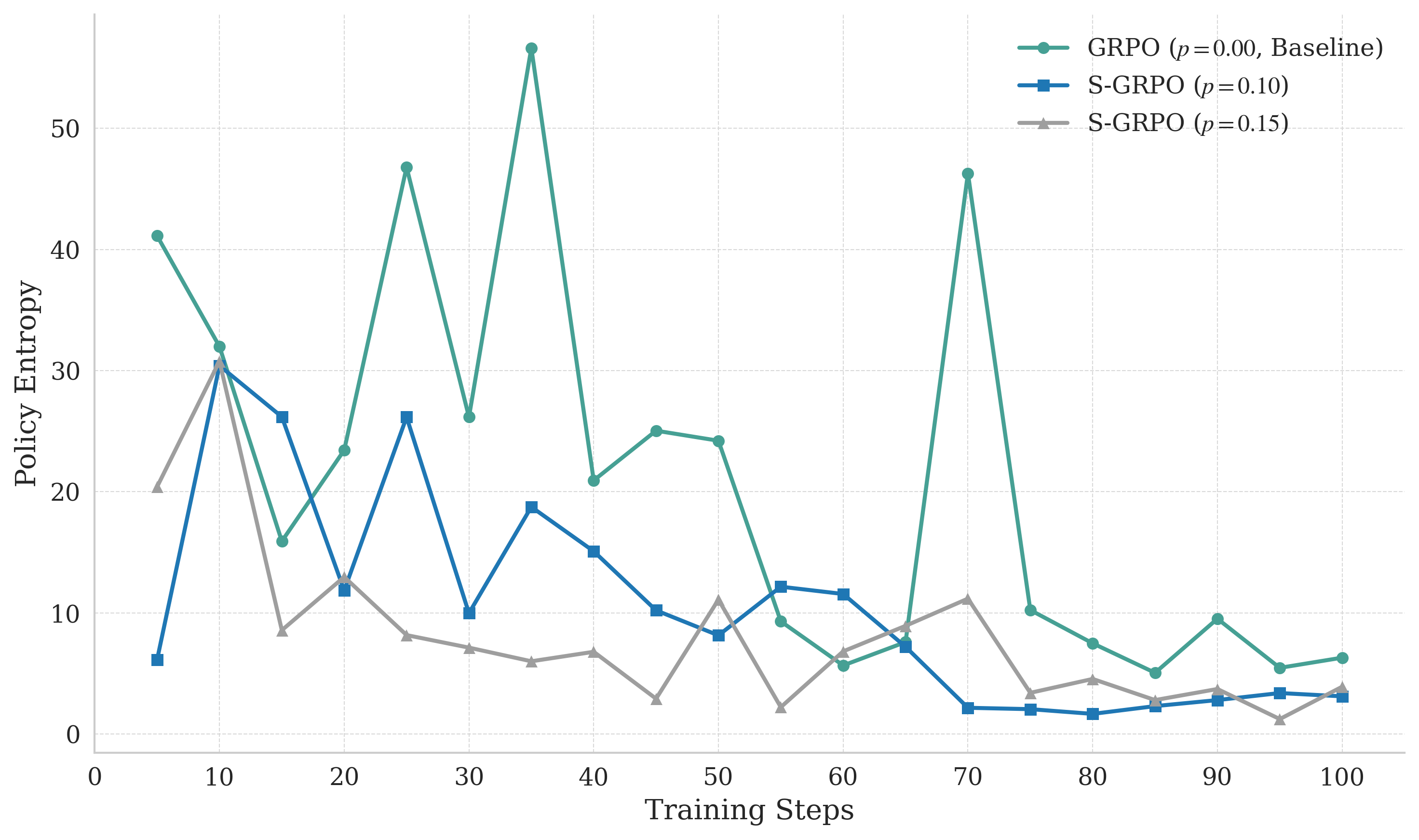}
	\caption{Policy entropy evolution under different noise assumptions on Qwen2.5-Math-7B-Base. Higher noise assumptions lead to smoother entropy reduction.}
	\label{fig:entropy}
\end{figure}

Recent work has identified the balance between exploration (policy entropy) and exploitation (accuracy) as crucial for RL-based reasoning models \citep{cheng2025reasoning,cui2025entropy}. Figure~\ref{fig:entropy} tracks entropy evolution during training, revealing how noise-aware reweighting affects exploration-exploitation balance.

Without reweighting ($p = 0$), policy entropy exhibits severe instability with dramatic fluctuations. This erratic behavior indicates the model alternates between near-deterministic policies and complete uncertainty, suggesting that conflicting signals from mismatched data cause repeated abandonment of learned behaviors.

In contrast, S-GRPO demonstrates controlled entropy reduction. At both $p = 0.10$ and $p = 0.15$, entropy decreases smoothly with only minor fluctuations. This monotonic decrease indicates consistent exploration throughout training, with gradual transition to exploitation.

\subsection{Analysis of S-GRPO Characteristics (RQ3)}

\subsubsection{Emergence of Self-Reflection Patterns}

\begin{figure}[ht]
	\centering
	\includegraphics[width=0.50\textwidth]{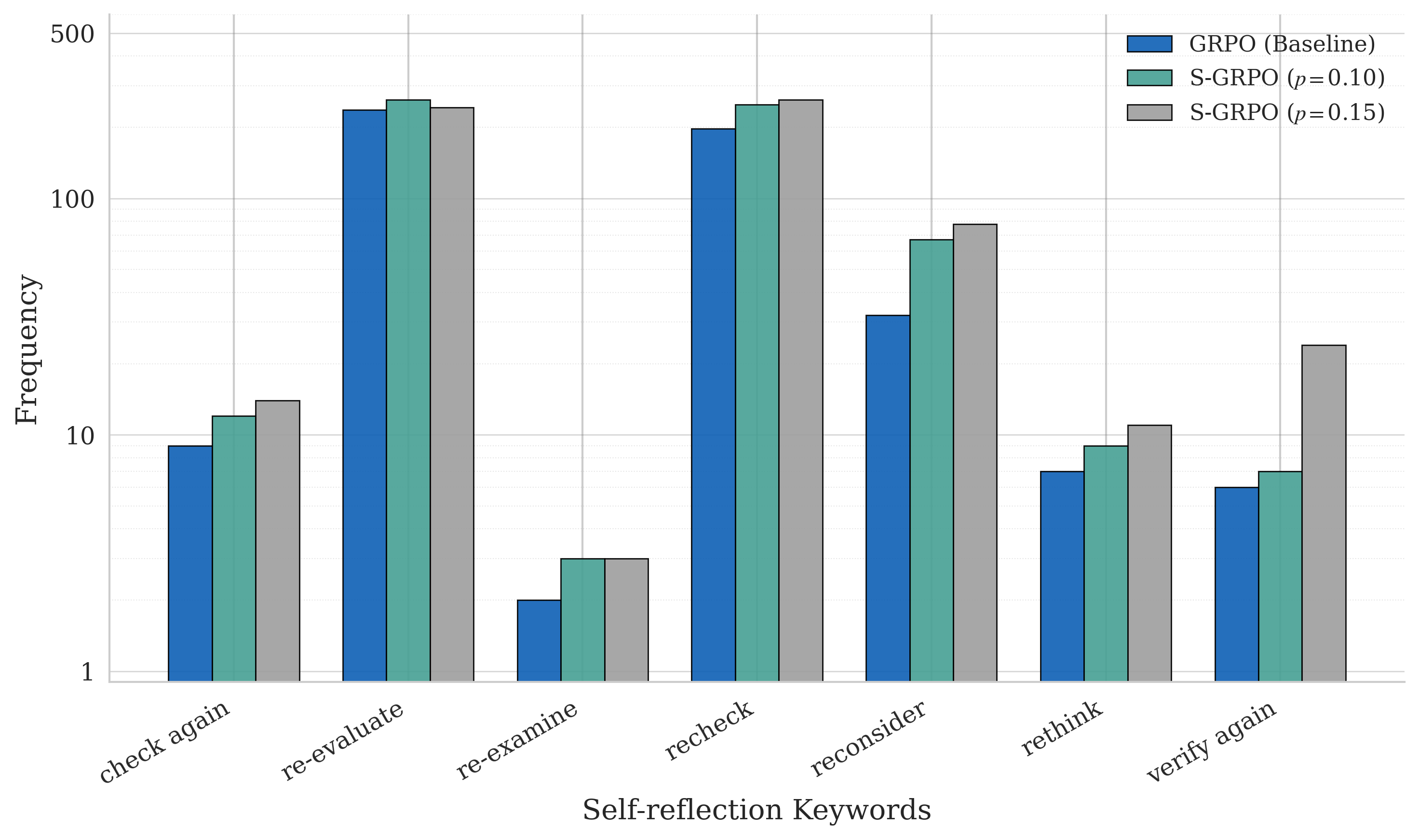}
	\caption{Frequency of self-reflection keywords in model responses. Higher noise assumptions correlate with increased self-reflection behavior across all keyword categories.}
	\label{fig:keywords}
\end{figure}

Self-reflection is considered a hallmark of effective reasoning \citep{guo2025deepseek,Min2024ImitateEA,Muennighoff2025s1ST}. Following \citet{liu2025understanding}, we analyze the frequency of self-reflection keywords in model outputs (Figure~\ref{fig:keywords}).

Results show that higher noise assumptions (larger $p$) consistently correlate with increased self-reflection behavior across all keyword categories. This suggests that noise-aware training facilitates the emergence of more sophisticated reasoning patterns by preventing premature convergence on superficial solution strategies.

\subsubsection{Response Length Analysis}

\begin{figure}[h]
	\centering
	\includegraphics[width=0.50\textwidth]{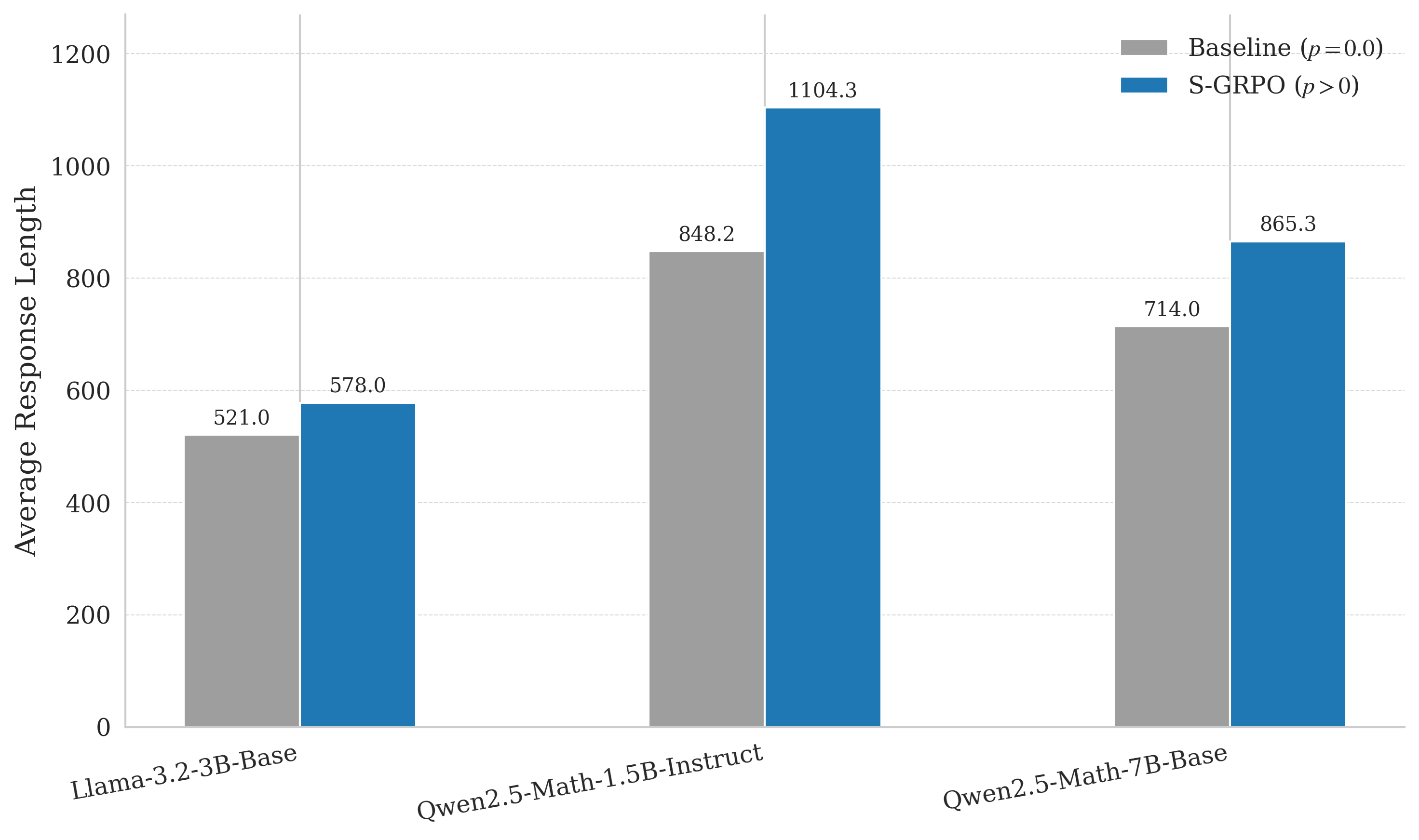}
	\caption{Comparison of response lengths for different models. S-GRPO consistently generates longer, more detailed responses across all base models.}
	\label{fig:length}
\end{figure}

Response length often correlates with reasoning depth \citep{Muennighoff2025s1ST,shen2025long}. Figure~\ref{fig:length} compares average response lengths across models and baselines. S-GRPO consistently generates longer responses than baselines across all base models, with improvements ranging from 13\% to 30\%.

\subsubsection{Ablation Study of Optimal Noise Assumptions}

\begin{figure}[h]
	\centering
	\includegraphics[width=0.45\textwidth]{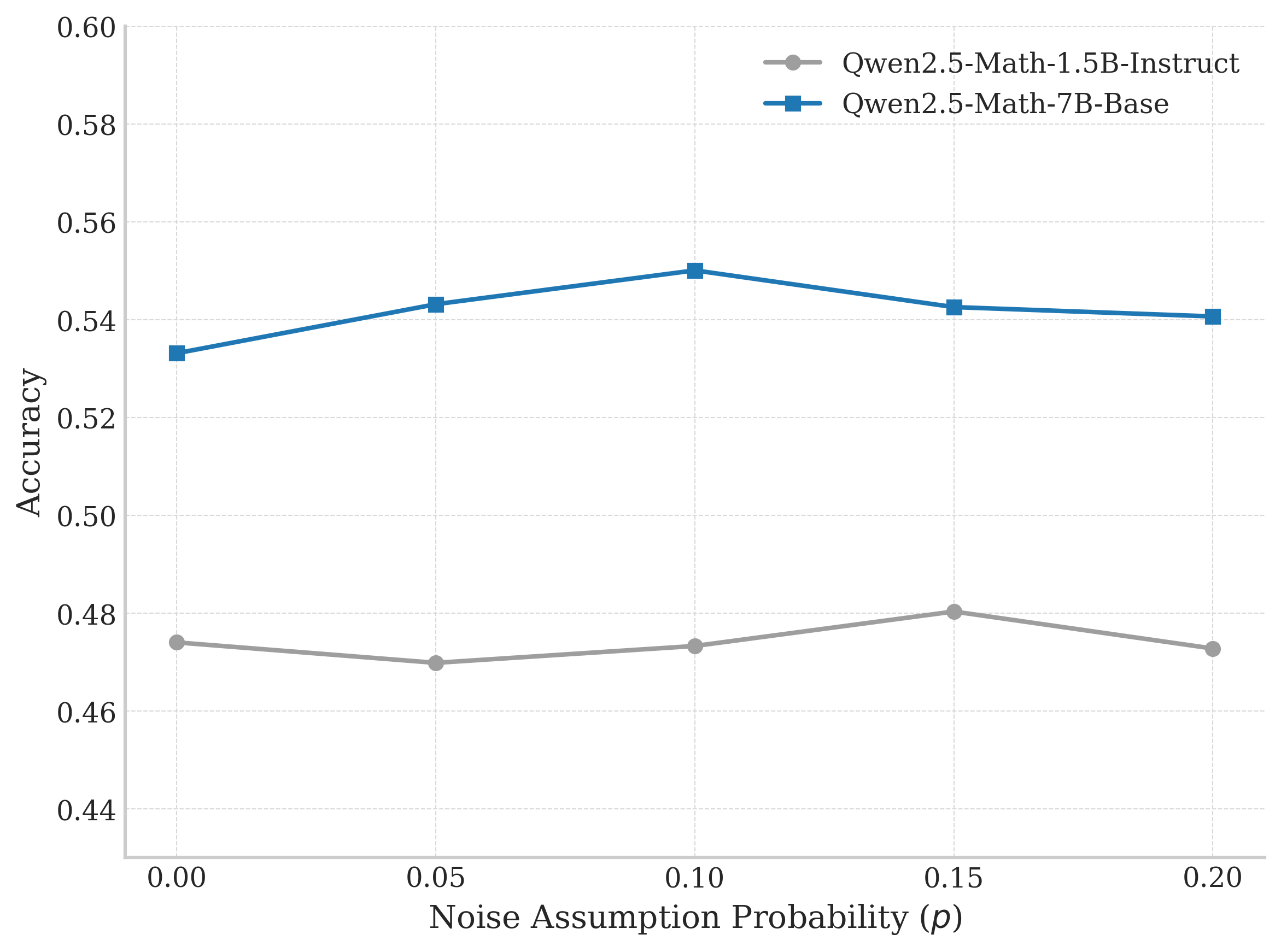}
	\caption{Effect of noise level $p$ on final performance. Optimal values differ by model scale, with larger models requiring lower noise assumptions.}
	\label{fig:ablation}
\end{figure}

Figure~\ref{fig:ablation} examines the effect of different noise assumption levels $p$ on final performance. Due to computational constraints, we limit training to 300 steps and report the average of top-3 checkpoints.

Our findings reveal model-dependent optimal noise levels: Qwen2.5-Math-1.5B-Instruct achieves best performance at $p = 0.15$, while the larger Qwen2.5-Math-7B-Base performs optimally at $p = 0.10$. This pattern indicates that larger models exhibit lower optimal noise levels, suggesting they naturally produce fewer think-answer mismatches during training.

The higher optimal $p = 0.15$ for the 1.5B model has important implications: it excludes extreme groups ($k \in \{1, 7\}$ out of 8 rollouts) from training updates. This aggressive filtering provides net benefits for weaker models, where the risk of corrupted learning signals from highly imbalanced groups outweighs the potential information loss from excluding these samples.

\subsubsection{Qualitative Analysis}

We analyze how S-GRPO's mitigation of Think-Answer Mismatch during training produces fundamentally different learned behaviors compared to baseline GRPO.

On a polynomial interpolation problem ($p(n) = \frac{n}{n^2-1}$ for $n=2,\dots,7$, find $p(8)$), the GRPO model outputs mathematical nonsense (a formula mixing unrelated concepts), which yields $\frac{8}{63}$. The S-GRPO model maintains coherent reasoning via auxiliary polynomial $q(x) = (x^2-1)p(x)-x$, though arriving at incorrect $\frac{5767}{63}$. This reveals how standard GRPO can learn to abandon logic when such gibberish accidentally succeeds during training.

For $(\sqrt{7}+\sqrt{5})^6$, GRPO attempts direct computation of $x^6+y^6$ beyond its capability and hallucinates "2916" (correct: 13536). S-GRPO instead decomposes systematically: $a^2 = 12+2\sqrt{35} \rightarrow a^4 = 284+48\sqrt{35} \rightarrow a^6 = 6768+1144\sqrt{35}$, maintaining context throughout.

These patterns demonstrate that S-GRPO's noise-aware reweighting during training filters out rewards from computational overreach and logical inconsistency, producing models that maintain reasoning coherence and respect computational boundaries. The details of examples are shown in Appendix B.
\section{Related Work}

\paragraph{GRPO and Variants.}
Group-Relative Policy Optimization (GRPO) \citep{shao2024deepseekmath} revolutionized LLM reasoning training by eliminating value functions and computing advantages relative to peer responses. There are two major lines of algorithmic improvements. The first focuses on controlling exploration behavior for more effective learning, such as clipping higher to mitigate entropy collapse \citep{yu2025dapo} or selectively updating only the 20\% of tokens with high entropy \citep{wang2025beyond}. These methods are orthogonal to ours and can be combined. The second line improves group-level reward computation, including Dr.~GRPO \citep{liu2025understanding}, SEED-GRPO \citep{chen2025seed}, and \citet{dang2025reinforcement}. However, these methods are primarily empirically motivated rather than grounded in noise theory. Our S-GRPO uniquely provides a principled, noise-aware reweighting solution derived from first principles.

\paragraph{Think-Answer Mismatch.}
The disconnect between correct answers and valid reasoning has been extensively documented \citep{lightman2023lets, chen2025reasoning}, with error rates ranging from 3.5\% to 51.8\% even when final answers are correct. While process supervision methods \citep{luo2024improve, lai2024step} aims to address this through expensive step-level annotations, our approach handles the mismatch at the outcome level through principled noise modeling, maintaining computational efficiency while improving robustness.

\paragraph{Noise-Robust Learning.}
Learning from noisy rewards has been addressed through various approaches including robust MDPs \citep{iyengar2005robust}, Bayesian methods \citep{yang2024bayesian}, and symmetric loss functions \citep{nishimori2025symmetric}. We adapt the classical symmetric noise model \citep{angluin1988learning} to derive closed-form optimal weights for GRPO. Unlike prior heuristic reweighting schemes, our approach is parameter-free given the noise estimate and requires no architectural changes or additional computation.

\section{Conclusion}

We identified and addressed a critical vulnerability in GRPO: its susceptibility to reward noise amplified by group imbalance. Our analysis revealed that a single mismatch sample can distort advantages by up to 60\% in unbalanced groups than in balanced groups, causing standard GRPO to fail entirely under a high noise levels.

S-GRPO provides a principled solution through noise-aware optimal reweighting. The derived weight $w^\star$ elegantly implements three key properties: noise-adaptive attenuation, confidence through consensus, and automatic gating of unreliable signals. These emerge naturally from minimizing expected squared error rather than heuristic design.

Empirically, S-GRPO achieves consistent 2-3\% improvements across diverse models while maintaining stable learning under 20\% noise where standard GRPO collapses. Beyond accuracy gains, S-GRPO fundamentally improves training dynamics, including promoting smooth entropy reduction, increased self-reflection, and more detailed reasoning.

This work demonstrates that principled analysis of algorithmic vulnerabilities can yield simple yet effective solutions. As reasoning models become critical infrastructure, such robustness improvements are essential for reliable deployment. Future work could explore asymmetric noise models, integration with process supervision, and applications to other RL algorithms for LLMs.


\acks{The authors declare that there is no funding support for this work. All authors have no competing interests related to this submission. All authors have read and approved the submission of this manuscript.}


\newpage

\appendix
\section{Experimental Setup}
\label{sec:appendix_setup}

Our implementation is built upon the public Open-source Algorithm-driven Training (OAT) framework\footnote{\url{https://github.com/sail-sg/oat}}. Our proposed S-GRPO method was implemented as a direct extension of the existing Dr.~GRPO module within the OAT codebase. This approach ensures that our comparisons to baselines are fair, as we inherit the vast majority of the underlying training pipeline and focus specifically on the impact of our noise-aware reweighting strategy.

All experiments were conducted on a cluster of 8 NVIDIA A100 (80GB) GPUs, using BF16 mixed-precision for training. A typical 500-step training run for a 7B model required approximately 40 hours. The key hyperparameters for our experiments are summarized in Table \ref{tab:hyperparams_main}.

\begin{table*}[h!]
	\centering
	\caption{Key hyperparameters for S-GRPO training runs.}
	\label{tab:hyperparams_main}
	\begin{tabular}{lc}
		\toprule
		\textbf{Parameter} & \textbf{Value} \\
		\midrule
		\multicolumn{2}{l}{\textit{Core Algorithm Parameters}} \\
		Base Framework & OAT \\
		Learning Rate & $1 \times 10^{-6}$ (Constant scheduler) \\
		S-GRPO Noise Level ($p$) & 0.10 (for 7B), 0.15 (for 1.5B/3B) \\
		Rollouts per Prompt ($N$) & 8 \\
		\midrule
		\multicolumn{2}{l}{\textit{Training \& Generation Configuration}} \\
		Global Training Batch Size & 128  \\
		Global Rollout Batch Size & 128 \\
		Max Generation Length & 4096 tokens \\
		Temperature & 1.0 \\
		Top-p & 0.95 \\
		Evaluation Frequency & Every 16 steps \\
		Base Models & Qwen2.5-Math-7B, Llama-3.2-3B, etc. \\
		Training Data & 8,500 problems from MATH (levels 3-5) \\
		Prompt Template & DeepSeek-R1 \\
		\bottomrule
	\end{tabular}
\end{table*}
\section{Case Study Analysis}

We provide detailed analysis of representative examples demonstrating how S-GRPO's mitigation of Think-Answer Mismatch during training leads to fundamentally different learned behaviors.

\subsection{Case 1: Coherence in Reasoning Framework}

\textbf{Problem:} Let $p(x)$ be a polynomial of degree 5 such that $p(n) = \frac{n}{n^2 - 1}$ for $n = 2, \dots, 7$. Find $p(8).$

\begin{tcolorbox}[sharp corners, boxrule=0.8pt, colback=white, fonttitle=\bfseries]
	\textbf{Model trained with GRPO:}
	\vspace{1mm}
	
	To find the polynomial \( p(x) \), we can use... However, a simpler approach is to use the polynomial:
	\textcolor{red}{\textbf{\[ p(x) = \frac{x}{(x^2 - 1) \frac{(x-2)(x-3)\dots(x-7)}{(1-2)(1-3)\dots(1-7)}}. \]}}
	...The value of \( p(8) \) is \(\boxed{\frac{8}{63}}\).
	
	\tcbline 
	
	\textbf{Model trained with S-GRPO:}
	\vspace{1mm}
	
	To solve for \( p(8) \), \textcolor{red}{\textbf{we can define a new polynomial \( q(x) = (x^2 - 1)p(x) - x \).}} This polynomial... has roots at \( x = 2, 3, \dots, 7 \). Since \( q(x) \) is a polynomial of degree 7, it can be written as:
	\textcolor{red}{\textbf{\[ q(x) = (x - 2)\dots(x - 7)(x - r) \]}}
	...The value of \( p(8) \) is \(\boxed{\frac{3}{56}}\).
\end{tcolorbox}

\textbf{Analysis:} This example illustrates how Think-Answer Mismatch corrupts GRPO training. The baseline model's formula is mathematical nonsense: polynomial interpolation concepts mixed with unrelated algebraic manipulations. Such outputs arise when training rewards correct final answers regardless of reasoning validity. During standard GRPO training, if such gibberish accidentally yields correct answers on some problems, the model learns that abandoning logical coherence can be a viable strategy.

S-GRPO's noise-aware reweighting mitigates this by down-weighting training signals from groups where such mismatches are likely. The resulting model demonstrates sophisticated mathematical reasoning: constructing the auxiliary polynomial $q(x)$ is a textbook approach for this problem type. This leads to a coherent reasoning framework, ultimately producing the correct result.

\subsection{Case 2: Fidelity in Computational Execution}

\textbf{Problem:} Find the greatest integer less than $(\sqrt{7} + \sqrt{5})^6$.

\begin{tcolorbox}[sharp corners, boxrule=0.8pt, colback=white, fonttitle=\bfseries]
	\textbf{Model trained with GRPO:}
	\vspace{1mm}
	...we start by letting \( x = \sqrt{7} + \sqrt{5} \) and \( y = \sqrt{7} - \sqrt{5} \). Notice that \( x^6 + y^6 \) is an integer...
	\vspace{1em}
	\textcolor{red}{\textbf{...By calculating the first few terms, we can see that \( x^6 + y^6 = 2916 \).}}
	\vspace{1em}
	...the greatest integer less than \( x^6 \) is \( 2915 \). The answer is \(\boxed{2915}\).
	
	\tcbline 
	
	\textbf{Model trained with S-GRPO:}
	\vspace{1mm}
	Let \(a = \sqrt{7} + \sqrt{5}\) and \(b = \sqrt{7} - \sqrt{5}\).
	\vspace{1mm}
	\textcolor{red}{\textbf{\begin{align*}
				a^2 &= 12 + 2\sqrt{35} \\
				a^4 &= (12 + 2\sqrt{35})^2 \\
				&= 284 + 48\sqrt{35} \\
				a^6 &= a^2 \cdot a^4 \\
				&= 6768 + 1144\sqrt{35} \\[0.5em]
				\Rightarrow a^6 + b^6 &= 13536
	\end{align*}}}
	...the greatest integer less than \(a^6\) is \(13535\). The answer is \(\boxed{13535}\).
\end{tcolorbox}

\textbf{Analysis:} This case reveals how Think-Answer Mismatch arises from insufficient context utilization when facing computational challenges. Both models correctly identify the conjugate approach, but their execution diverges dramatically.

The GRPO-trained model attempts to directly compute $x^6 + y^6$, which is a calculation requiring tracking binomial expansions with terms like $\binom{6}{k}(\sqrt{7})^k(\sqrt{5})^{6-k}$ that exceeds its reliable computational horizon. Unable to maintain context through this complex calculation, it resorts to hallucinating a value (2916 vs. correct 13536). This behavior emerges from training that rewards such computational overreach when lucky guesses yield correct final answers.

In contrast, the S-GRPO-trained model demonstrates learned respect for its computational boundaries through systematic decomposition. Each intermediate step ($a^2$, $a^4$, $a^6$) involves only a single algebraic operation within the model's capability, with explicit context propagation between steps. This incremental approach directly results from S-GRPO's training process: by down-weighting signals from unbalanced groups where computational overreach might accidentally succeed, it reinforces policies that favor verifiable, step-wise progress over high-risk leaps.

\bibliography{sample}

\end{document}